\documentclass[letterpaper, 10 pt, conference]{ieeeconf}  

\IEEEoverridecommandlockouts                              

\usepackage{amsmath} 
\usepackage{amssymb}  

\usepackage{todonotes}
\usepackage{subcaption}
\usepackage{tabularx}

\title{\LARGE \bf
Situational Adaptive Motion Prediction for Firefighting Squads in Indoor Search and Rescue
}

\author{Nils Mandischer$^{1}$, Frederik Schicks$^{1}$, and Burkhard Corves$^{1}$
\thanks{$^{1}$The authors are with the Institute of Mechanism Theory, Machine Dynamics and Robotics, RWTH Aachen University, 52062 Aachen, Germany
	Contact author: {\tt\small mandischer@igmr.rwth-aachen.de}}%
}

\begin{document}

\maketitle
\thispagestyle{empty}
\pagestyle{empty}


\begin{abstract}
	
	Firefighting is a complex, yet low automated task. To mitigate ergonomic and safety related risks on the human operators, robots could be deployed in a collaborative approach. To allow human-robot teams in firefighting, important basics are missing. Amongst other aspects, the robot must predict the human motion as occlusion is ever-present. In this work, we propose a novel motion prediction pipeline for firefighters' squads in indoor search and rescue. The squad paths are generated with an optimal graph-based planning approach representing firefighters' tactics. Paths are generated per room which allows to dynamically adapt the path locally without global re-planning. The motion of singular agents is simulated using a modification of the headed social force model. We evaluate the pipeline for feasibility with a novel data set generated from real footage and show the computational efficiency.
	
\end{abstract}

\section{Introduction}
\label{sec:introduction}
Indoor search and rescue (SAR) is exceptionally stressful to the firefighting operators. We aim to deploy humans and mobile robots as a collaborative squad to reduce health risks of diverse nature. A collaborative squad consists of multiple firefighting operators (the squad) and a collaborative rescue robot. Similar to the human, the robot is challenged by obscured vision on the environment and that conditions may change drastically throughout each mission. Therefore, the robot must deploy multiple methodologies to perceive and predict the location of its assigned human squad. This is analogous to approaches established in service robotics, where the robot perceives the human locations through sensors but is also able to predict their future whereabouts through motion modeling. For indoor SAR missions, no motion and behavioral models are known. In this work, we propose an approach to model human behavior throughout such missions by combining a graph-based tactics-informed optimal planning approach and a modification of the headed social force model (HSFM)~\cite{FFG17}. Optimal paths are planned according to common tactics considering restricted and unrestricted vision per room and per squad. The planning is based on a priori knowledge of the environment map and status, which are usually available at the start of a mission. Through the segmentation into room entities, paths are adapted to the current scene when predicted conditions change. Individual agents (i.e., the simulated firefighting operators), then, use the optimal paths as waypoints in their individual motion prediction. By this, inter- and intra-squad behavior is modeled. The models are parameterized according to real data and evaluated for usability in real-time systems.

\pagebreak
Our main contributions in modeling indoor SAR are:
\begin{itemize}
	\item New model and novel pipeline for motion prediction
	\item Composed novel data set to parameterize motion models
\end{itemize}

\section{Related Work}
Modeling of human behavior is a common task outside of SAR. Of particular interest for this work are physics based models with dynamic environment and group cues. One of the earliest adoptions is the social force model (SFM) by Helbing et al.~\cite{HM95,HFV00}, which was later adapted by many works: Moussa\"{i}d et al.~\cite{MPG10} add group cues to the SFM. Pellegrini et al.~\cite{PES09} introduce collision prediction. Yamaguchi et al.~\cite{YBO11} decide on the (hidden) group status by observing motion. Rudenko et al.~\cite{RPL18,RPA18} combine a velocity model based on a Markov decision process with SFM-based local interactions and a stochastic random walk policy.

Farina et al.~\cite{FFG17} extend the SFM to the HSFM by applying a locomotion model which prefers forward motion in gazing direction of the agent. The locomotion model is defined by
\begin{subequations}
	\begin{gather}
		\mathbf{R}(\theta_{i}) =
		\begin{bmatrix}
			\mathbf{e}_{x} & \mathbf{e}_{y}
		\end{bmatrix} =
		\begin{bmatrix}
			\cos{\theta_{i}} & -\sin{\theta_{i}}\\
			\sin{\theta_{i}} & \cos{\theta_{i}}
		\end{bmatrix},\\
		\dot{\mathbf{p}}_{i} = \mathbf{R}(\theta_{i})\mathbf{v}_{i}, \quad
		\dot{\mathbf{v}}_{i} = \frac{1}{m_{i}}\mathbf{u}_i^{B}, \quad
		\dot{\omega}_{i} = \frac{1}{I_{i}}u_{i}^{\theta}, 
		\tag{\addtocounter{equation}{1}\theequation,\addtocounter{equation}{1}\theequation,\addtocounter{equation}{1}\theequation}
	\end{gather}
\end{subequations}
where an agent $i$ is characterized by their pose \mbox{$\mathbf{q}_{i}=\begin{bmatrix}\mathbf{p}_{i} & \theta_{i}\end{bmatrix}^{T}$}, velocity \mbox{$\dot{\mathbf{q}}_{i}=\begin{bmatrix}\dot{\mathbf{p}}_{i} & \omega_{i}\end{bmatrix}^{T}$}, mass $m_{i}$, and inertia $I_{i}$, accordingly. The gazing direction $\theta_{i}$ is spanned between the global frame and a frame co-moving with the agent, hence, the distinction of velocities into $\dot{\mathbf{p}}_{i}$ (global) and $\mathbf{v}_{i}=\begin{bmatrix} v_{i,x} & v_{i,y} \end{bmatrix}^{T}$ (local). The lateral input forces $\mathbf{u}_{i}^{B}=\begin{bmatrix}u_{i}^{f} & u_{i}^{o}\end{bmatrix}^{T}$ and torque input $u_{i}^{\theta}$ are deconstructions of the social forces, defined by
\begin{subequations}
	\begin{align}
		u_{i}^{f} &= \mathbf{f}_{i}\mathbf{e}_{x}(\theta_{i}),\\
		u_{i}^{o} &= C^{o}(\mathbf{f}_{i}-\mathbf{f}_{i}^{acc})\mathbf{e}_{y}(\theta_{i})-C^{des}v_{i}^{des,o},\\
		u_{i}^{\theta} &= -C^{\theta}(\theta_{i}-\phi_{i}^{acc})-C^{\omega}\omega_{i},
	\end{align}
\end{subequations}
with the social forces resulting from external factors, i.e., agent-agent $\mathbf{f}_{i}^{j}$ and agent-border $\mathbf{f}_{i}^{B,k}$ interaction, and the agent's motivation to reach the goal $\mathbf{f}_{i}^{acc}$. The total social force $\mathbf{f}_{i}$ acting on the individual agent $i$ is the sum of all partial forces, defined by
\begin{equation}
	\mathbf{f}_{i} = \mathbf{f}_{i}^{acc} + \sum_{j\ne i}^{\text{agents}}\mathbf{f}_{i}^{j} + \sum_{k}^{\text{borders}}\mathbf{f}_{i}^{B,k}.
\end{equation}
In this work, $\phi_{i}^{acc}$ is the phase of $\mathbf{f}_{i}^{acc}=\{d_{i}^{acc},\phi_{i}^{acc}\}$. $C^{o}=1$, $C^{des}=500$ are scaling parameters and $v_{i}^{des,o}$ is the desired velocity in orthogonal direction. The torque parameters $C^{\theta}$ and $C^{\omega}$ are configured according to~\cite{FFG17}.

\section{Tactics and Restrictions of the Model}
\label{sec:tactics}
Fire brigades worldwide apply different tactics, however, the overall approach is similar. A mission is usually split into multiple stages, e.g., the German THW employs five~\cite{MMG14} and the Australian NDO six stages~\cite{Dep89}. This work relates to firefighting in Germany, particularly, to the attack stage, which involves exploring indoor disaster environments, commonly, with respiratory protection~\cite{CAL04}.

\subsection{Motion Tactics in Indoor Attack}
\begin{figure}[t!]
	\centering
	\begin{subfigure}[h]{.15\textwidth}
		\centering
		\includegraphics[width=\textwidth]{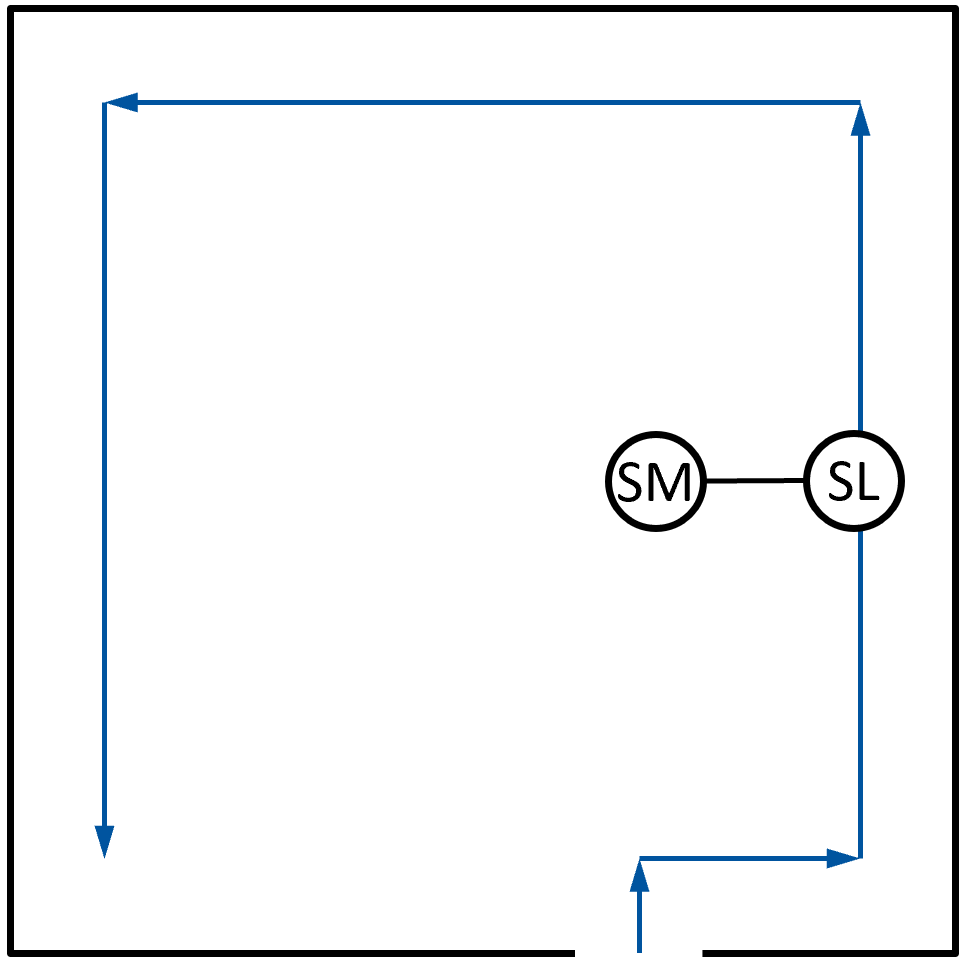}
		\caption{Wall search.}
		\label{sfig:goalgeneration_search_tactics_wall}
	\end{subfigure}
	\hfill
	\begin{subfigure}[h]{.15\textwidth}
		\centering
		\includegraphics[width=\textwidth]{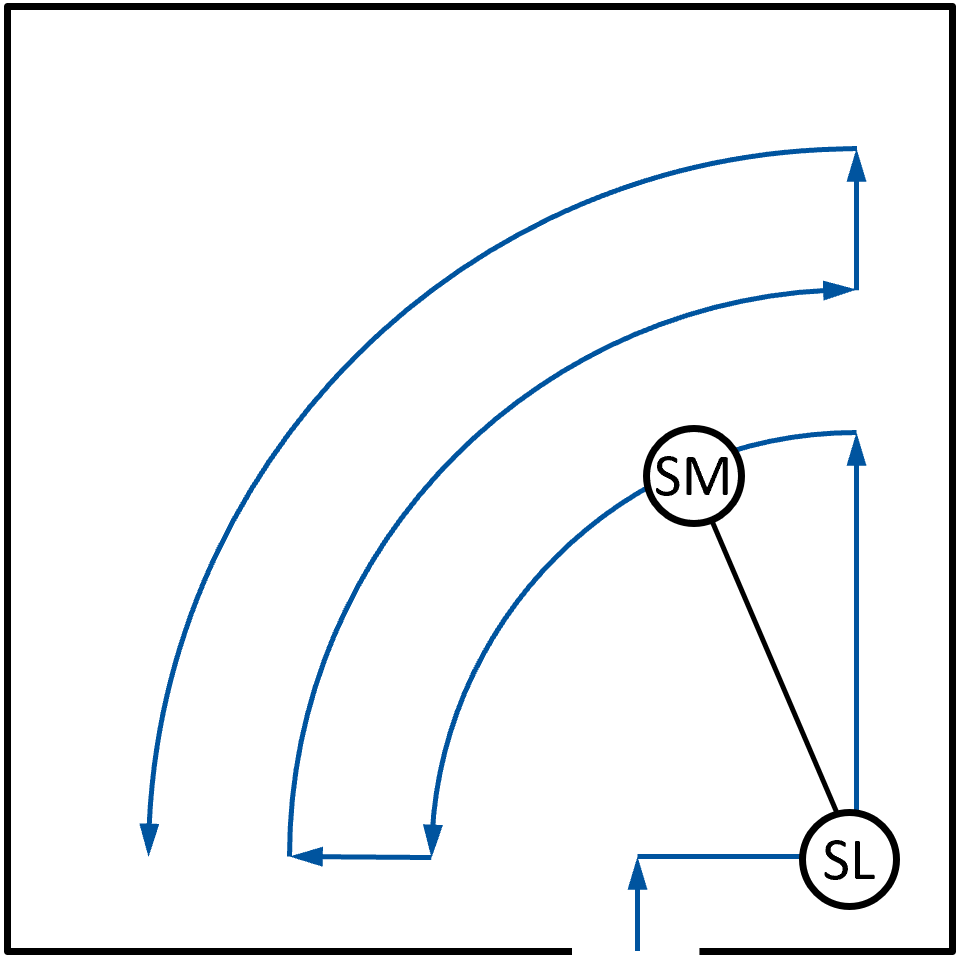}
		\caption{Diving search.}
		\label{sfig:goalgeneration_search_tactics_diving}
	\end{subfigure}
	\hfill
	\begin{subfigure}[h]{.15\textwidth}
		\centering
		\includegraphics[width=\textwidth]{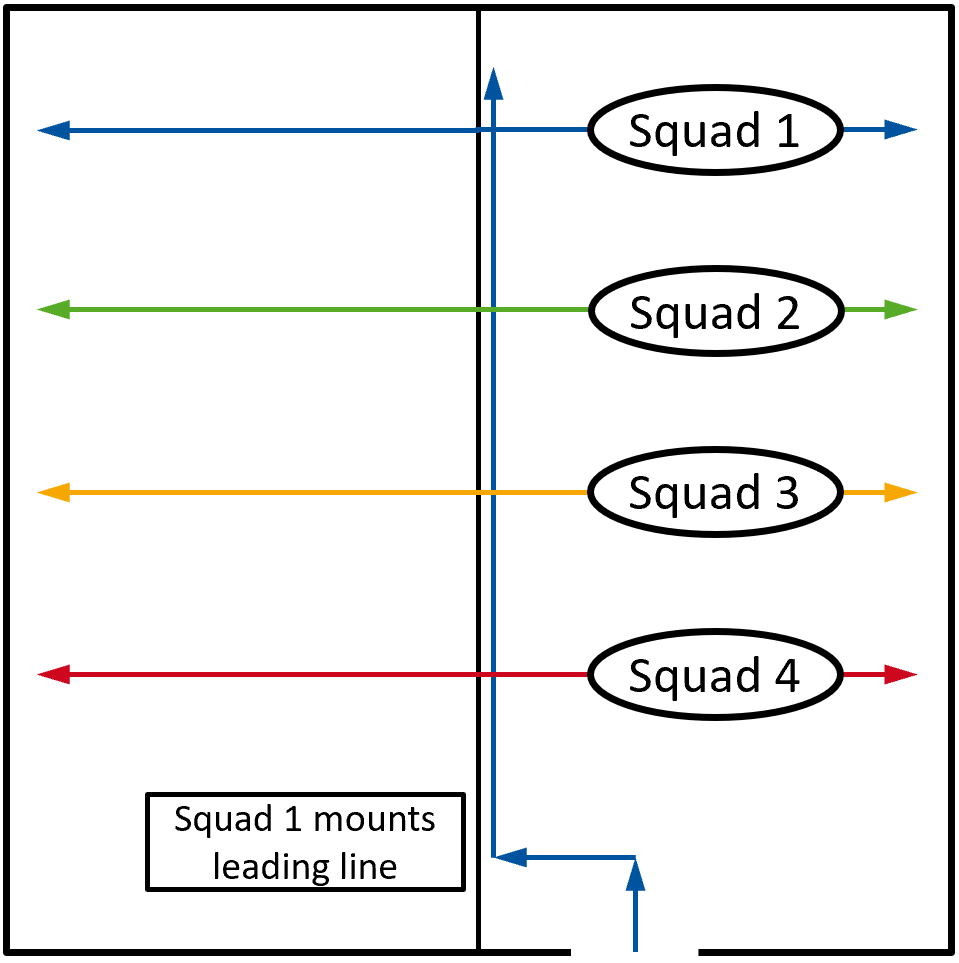}
		\caption{Tree search.}
		\label{sfig:goalgeneration_search_tactics_tree}
	\end{subfigure}
	\setlength{\belowcaptionskip}{-15pt}
	\caption{Search tactics under restricted vision in individual rooms. Wall and diving search use base tactic RHR to approach the room (SL: squad leader, SM: squad mate).}
	\label{fig:goalgeneration_search_tactics}
\end{figure}

During indoor attack, motion tactics are applied per room based on its size and the visual conditions. In Germany, four different tactics are applied. In non-obscured vision, the squad explores the room until the entire space has been visually inspected without further instructions on exact motion (referred to as free traversal). In case of vision restriction, wall search (from German: ``Wandtechnik'', Figure~\ref{sfig:goalgeneration_search_tactics_wall}) is applied. Within, one firefighter constantly keeps their hand to the right- or left-hand wall, hence, the distinction in right-hand rule (RHR) and left-hand rule (LHR), respectively. They stretch out their arms and utilize squad mates and equipment to extend their reach. This technique is used in small to medium sized rooms, which are common in civil buildings, e.g., dorms or office buildings. In case of larger rooms, diving search (``Tauchertechnik'', Figure~\ref{sfig:goalgeneration_search_tactics_diving}) and tree search (``Baumtechnik'', Figure~\ref{sfig:goalgeneration_search_tactics_tree}) are applied. To establish a first baseline for firefighters' motion prediction, we focus on smaller rooms, hence, only free traversal and wall search (RHR and LRH) are modeled in this work.

\subsection{Motion Data}

For motion modeling, not only pure knowledge about the tactics, but also data is required to parameterize the models. However, real-world data is hard to come by as firefighters usually do not carry recording devices. Therefore, we base our data set on a thermal recording of the \emph{Institut der Feuerwehr NRW} (IdF) generated in the project \emph{KOORDINATOR}~\cite{BDW14}, which depicts approaches under vision restriction with wall search, and two TV shows~\cite{BK17,RM17}. The TV shows display real footage of indoor SAR with and without vision restriction. Of particular interest are the velocities and the distances the squad applies to borders and between squad members. The parameters we tune our model to are listed in Table~\ref{tab:data}. All parameters have high variance due to the low sample size. Particularly, velocity data is not trustworthy. However, as there is virtually no data for indoor SAR, these statistics implement vital reference values.


\section{Graph-Based Tactics-Informed Optimal Planning}
\label{ssec:framework_graph}
The prediction framework is split into two stages. First, waypoints are generated by optimal path planning per room and squad. These are, then, used in the HSFM to generate agent-individual trajectories. The framework can generate motion trajectories of single agents as well as of individual agents in a single or in multiple squads. As input, a building plan is required, which is usually available for newer buildings and may be interpreted by other means (compare~\cite{PHS22} for an overview). The map is segmented into individual rooms and their connecting doorways. These are, then, stored in a room graph in which nodes are rooms and vertices are doorways between them. For each room, a graph is generated which expands the room's node as a sub-graph embedded in the room graph. By this, global room sequences and local paths in each room may be planned isolated or together in a global optimization approach.

\begin{table}
	\caption{Parameters derived from \cite{BDW14,BK17,RM17}.}
	\centering
	\begin{tabular}{r r | c c c}
		\hline
		\textbf{Parameter} & \textbf{Vision} & \textbf{Mean ($\mu$)} & \textbf{Variance ($s^2$)} & \textbf{Samples}\\
		\hline
		Dist. in squad  & Free   & $0.634$ & $0.301$ & $105$\\
		(in m)          & Restr. & $0.275$ & $0.073$ & $325$\\
		\hline
		Dist. to border & Free   & $0.392$ & $0.106$ & $76$\\
		(in m)          & Restr. & $0.297$ & $0.134$ & $188$\\
		\hline
		Velocity        & Free   & $1.500$ & $0.212$ & $22$\\
		(in m/s)        & Restr. & $0.326$ & $0.056$ & $29$\\
		\hline
	\end{tabular}
	\label{tab:data}
\end{table}

\subsection{Graph Construction}
\begin{figure}[t!]
	\centering
	\begin{subfigure}[t]{.23\textwidth}
		\centering
		\includegraphics[width=\textwidth]{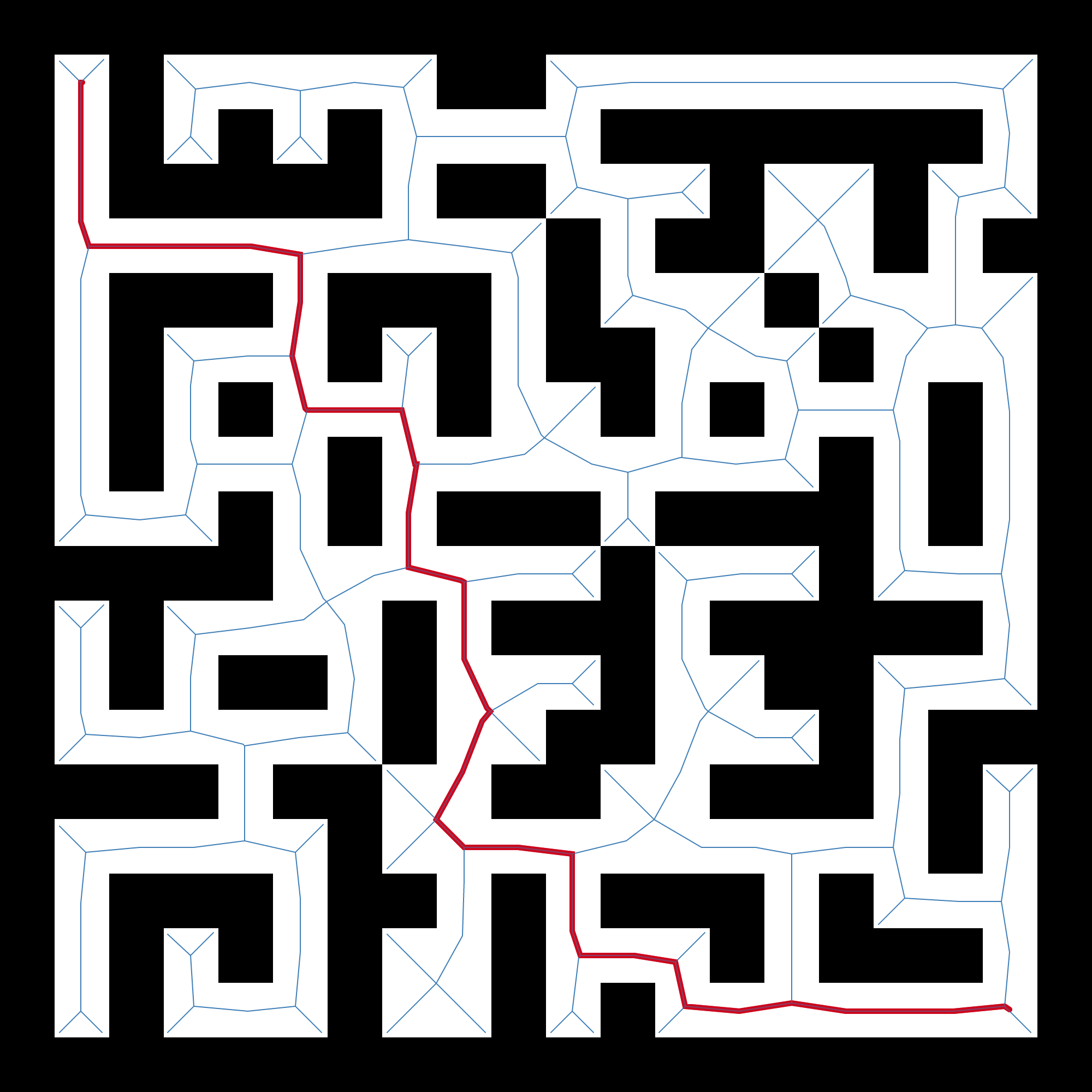}
		\caption{Medial axis.}
		\label{sfig:graph_medial}
	\end{subfigure}
	\hfill
	\begin{subfigure}[t]{.23\textwidth}
		\centering
		\includegraphics[width=\textwidth]{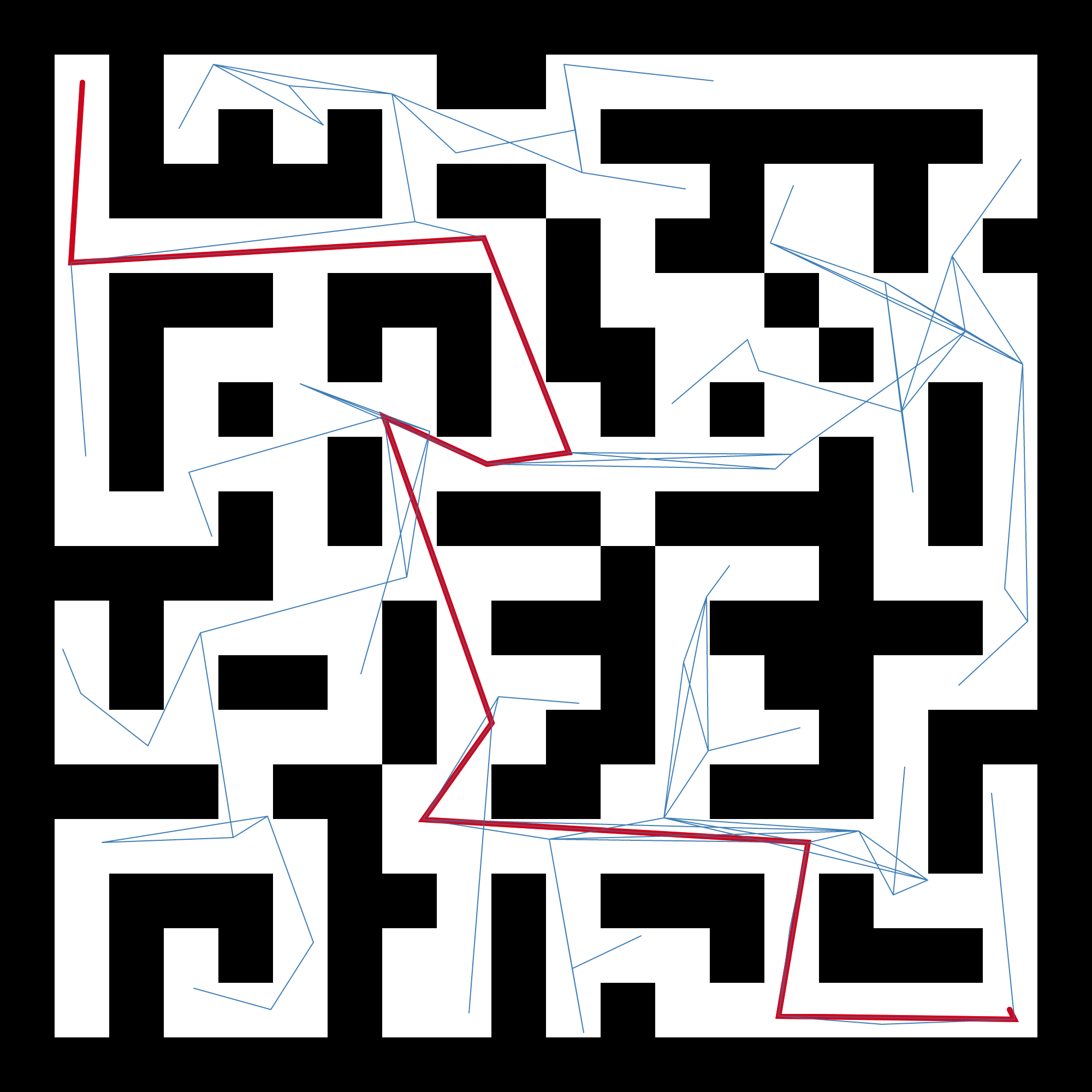}
		\caption{Visibility road map.}
		\label{sfig:graph_visibility}
	\end{subfigure}
	\vfill
	\begin{subfigure}[t]{.23\textwidth}
		\centering
		\includegraphics[width=\textwidth]{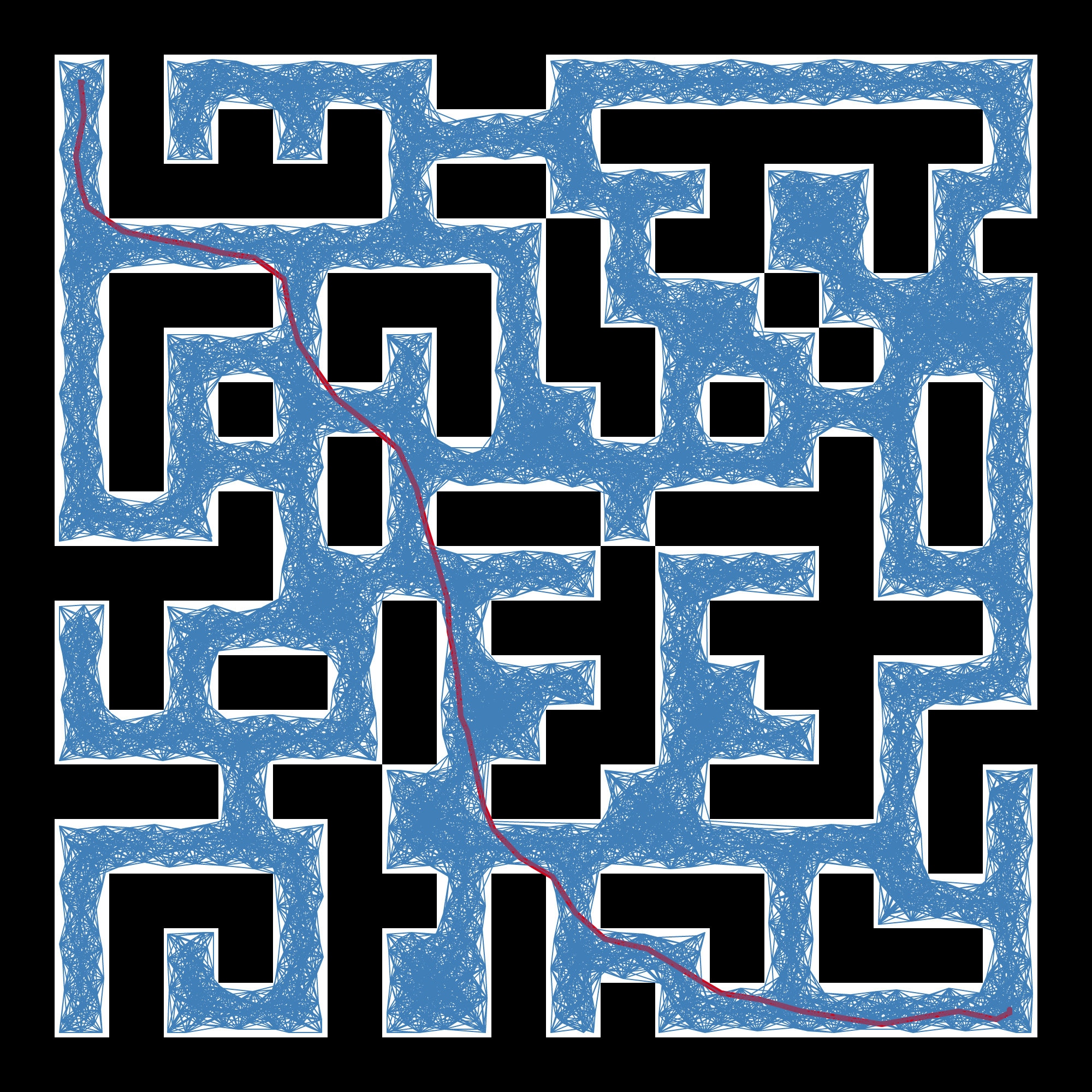}
		\caption{Combined graph of \subref{sfig:graph_medial}, \subref{sfig:graph_visibility}, and pseudo-random sampling.}
		\label{sfig:graph_combined}
	\end{subfigure}
	\hskip.018\textwidth
	\begin{subfigure}[t]{.23\textwidth}
		\centering
		\includegraphics[width=\textwidth]{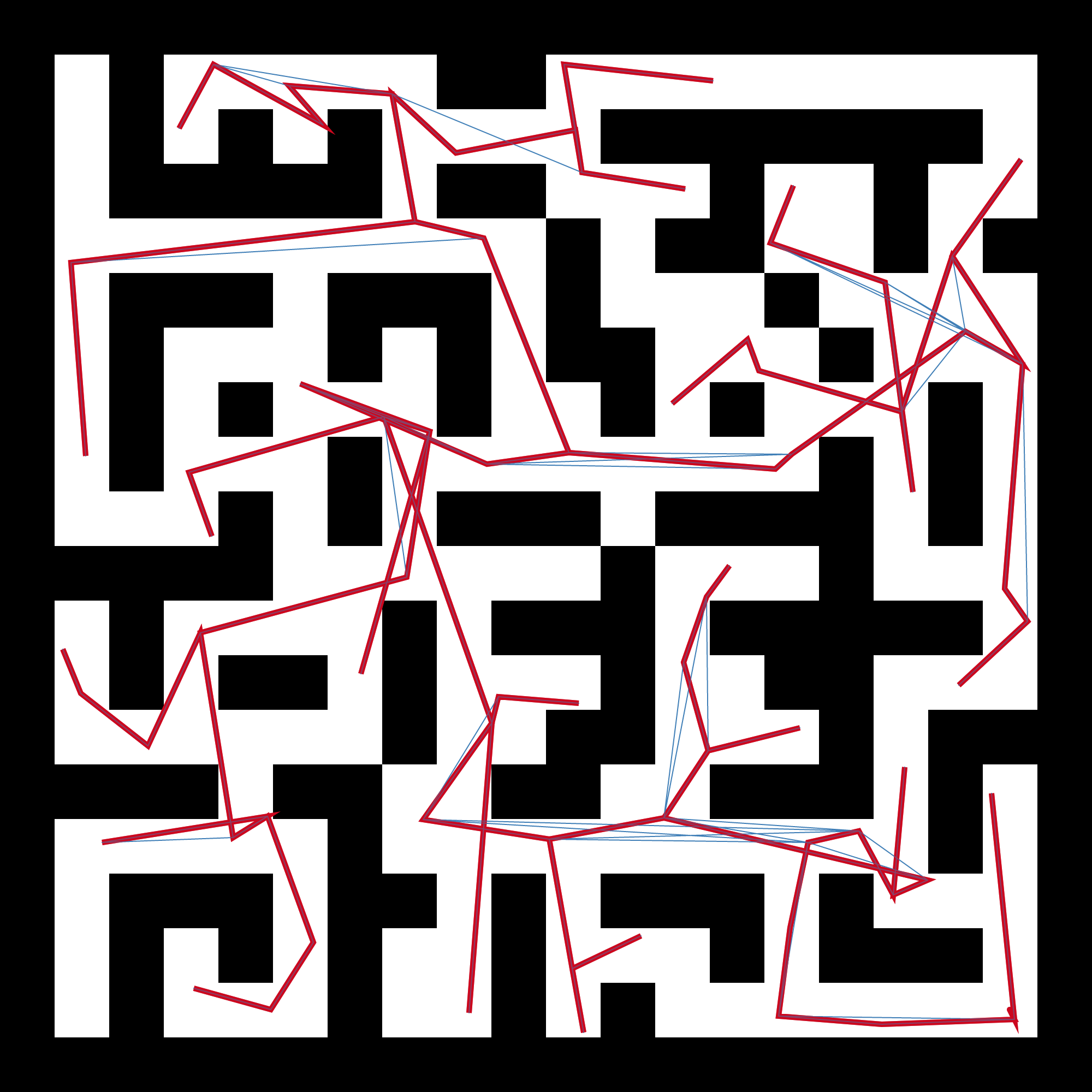}
		\caption{Greedy tree search on~\subref{sfig:graph_visibility}.}
		\label{sfig:graph_tree}
	\end{subfigure}
	\setlength{\belowcaptionskip}{-15pt}
	\caption{Graphs generated (blue) and path planned (red) in free traversal on complex artificial map.}
	\label{fig:graphs}
\end{figure}

For graph construction, we use three different types of graphs: Medial axis, visibility road map and pseudo-random sampling. The \textbf{medial axis} (Figure~\ref{sfig:graph_medial}) is constructed from a Voronoi graph using the method proposed by Masehian and Amin-Naseri~\cite{MA04}. The seeds of the Voronoi regions are located in the occupied regions of the grid map. These are determined using distance transforms~\cite{Bor86}. The medial axis is a graph that is constructed only from Voronoi border regions, i.e., where multiple Voronoi regions meet. Hence, it generates a graph with maximal distance to boundaries. The \textbf{visibility road map}~\cite{NSL99} (Figure~\ref{sfig:graph_visibility}) generates a graph which guarantees that all space is visible. During construction, nodes are sampled pseudo-randomly using the Halton sequence~\cite{Hal60}. Each sampled node is checked for visibility and added if it (a) is not visible from any other node (called guard) or (b) connects at least two guard nodes that were not connected beforehand (called connector). By checking the visibility road map it is determined whether space has been observed, which is particularly important when modeling free traversal behavior in SAR. Firefighters explore rooms until the entire space has been inspected to locate casualties. The remaining space is filled with nodes \textbf{pseudo-randomly sampled} with the Hammersley sequence~\cite{Ham64}.


Nodes are connected based on proximity, i.e., all neighbors in a specified radius are connected by vertices (Figure~\ref{sfig:graph_combined}). The vertices are generated bidirectional, whereas each directed vertex carries information on the Euclidean length and in which motion tactics it is permissible for path planning. Free traversal allows all vertices. However, wall search requires locations close to walls, hence, vertices to far regions are blocked. Further, depending on the direction of motion, only clockwise (LHR) or counter-clockwise (RHR) vertices are allowed. On the combined graph, the shortest path within the given vertex restrictions is chosen. We observe good path quality with the A*~algorithm in all three types of motion.

\subsection{Special Case: Single Entry Rooms}
In rooms with a single entry, the agent would be allowed to directly move back out, given the prior explanation. This is prevented by following considerations: \underline{Free traversal} additionally requires all nodes in the visibility road map to be visited. To guarantee this, we build a tree from the entry node and follow the closest vertices in a greedy manner until a leaf is reached (Figure~\ref{sfig:graph_tree}). Then new branches are generated from previously visited nodes until a new leaf is reached. In \underline{wall search}, the first node in the path is moved by one margin clockwise (LHR) or counter-clockwise (RHR). As backtracking vertices are blocked, the agent is forced to move along the wall through the room instead of moving out of the doorway.

\section{Modified Headed Social Force Model}
\label{ssec:framework_hsfm}
After paths have been generated per squad, they are propagated to each agent in their respective squad. Motion trajectories are simulated using a modification of the HSFM.

\subsection{Waypoint Management}
\label{ssec:waypoint_management}
An agent derives the full list of waypoints from its squad's path. Agents delete their waypoints individually if latter have been visited. A waypoint is marked visited if it is located (a) within a cone in the agent's gazing direction or (b) within a slim circle centered at the agent's position. The cone has a range of $50$m in free and $2$m in restricted vision, and an opening angle of $180$°. The circle is defined with a range of $0.2$m. Latter's main purpose is to delete waypoints if an agent is spawned exactly on top, in which case the cone alone is not sufficient. Waypoints are marked essential if they are located at key locations, e.g., doorways, or start/end points. An essential waypoint is only deleted using criterion (b), but with a relaxed threshold. Thus, agents maintain a goal over long distances which prevents them from sliding along walls.

\subsection{Modified Contact Model}
In the HSMF~\cite{FFG17}, agents typically stay far from walls, e.g., in corridors they tend to walk in the middle. In firefighting, however, typical behavior (particularly in wall search) leads to agents staying much closer to walls. In addition, to safe computational resources our aim is to renounce the usage of high level extraction of semantic building entities besides rooms, e.g., the agglomeration of occupied pixels into walls. These two aspects lead to the challenge, that agents may get pushed through walls and once they are, there is no way back, as it is not possible to determine from which side an occupied area was penetrated. Note that no past data is stored.
\begin{figure}[t!]
	\centering
	\begin{subfigure}[b]{.23\textwidth}
		\centering
		\includegraphics[width=\textwidth]{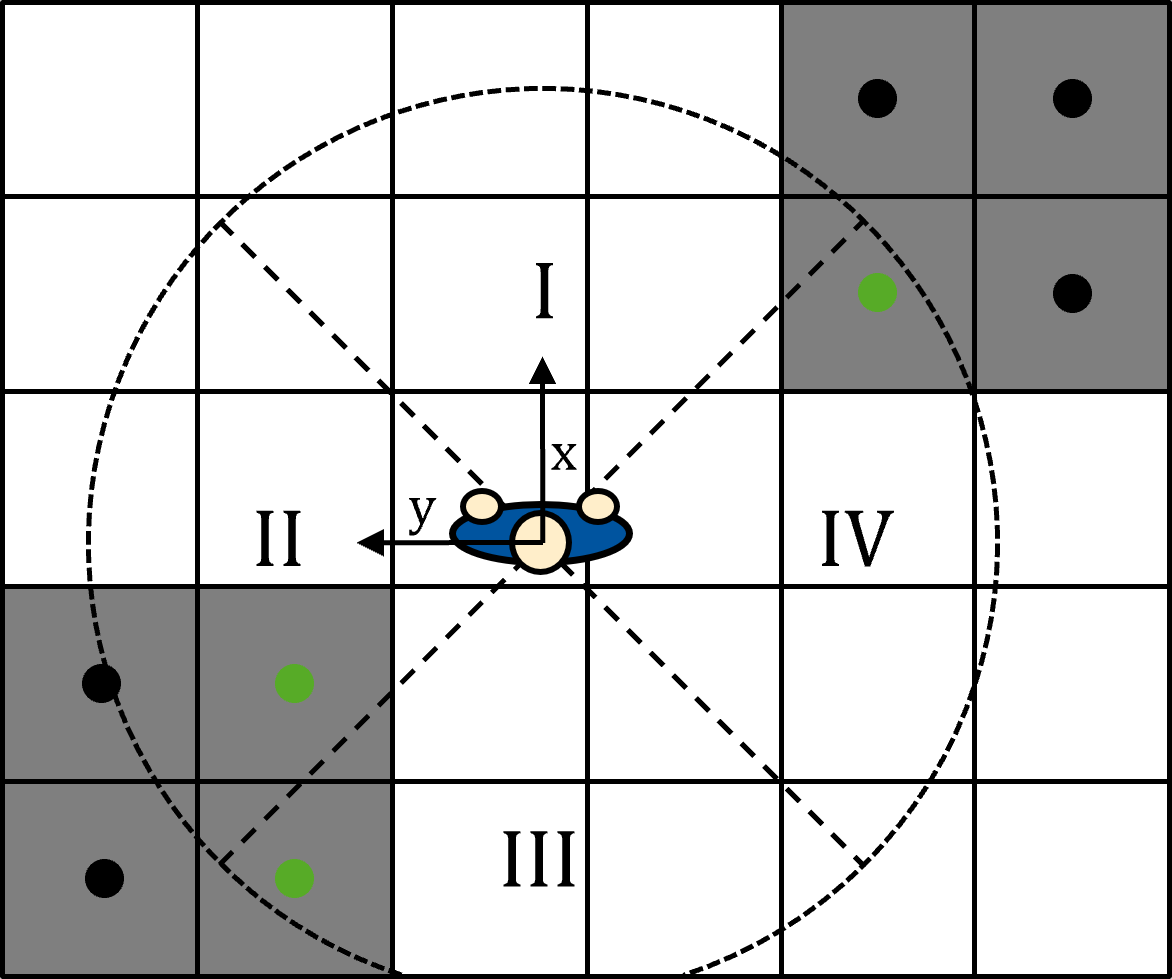}
		\caption{Four quadrants and occupied space considered for border forces (green).}
		\label{sfig:contact_quadrants}
	\end{subfigure}
	\hfill
	\begin{subfigure}[b]{.23\textwidth}
		\centering
		\includegraphics[width=\textwidth]{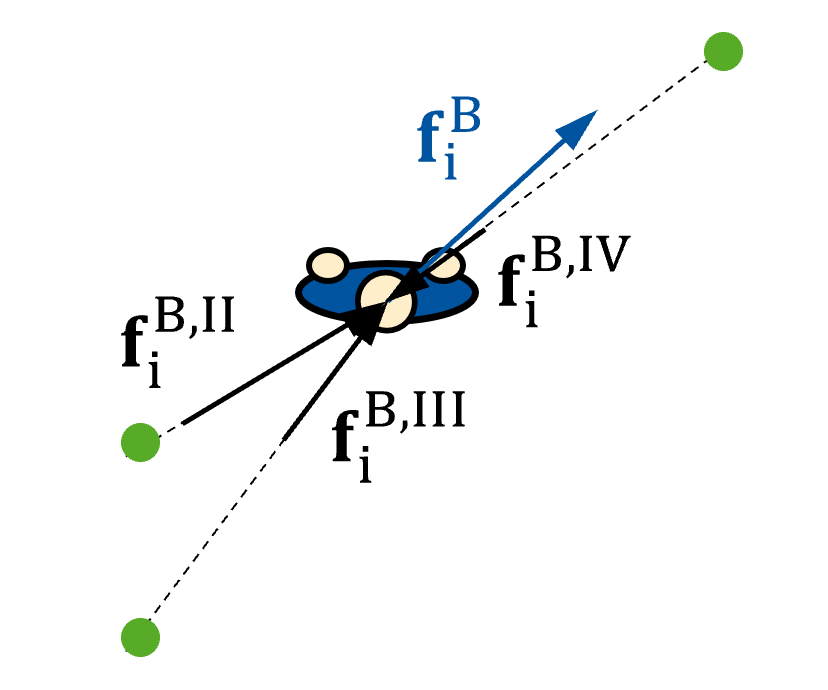}
		\caption{Partial forces $\mathbf{f}_{i}^{B,k}$ and resulting force $\mathbf{f}_{i}^{B}$ according to considered centers of mass.}
		\label{sfig:contact_forces}
	\end{subfigure}
	\setlength{\belowcaptionskip}{-15pt}
	\caption{Sketch of the agent-border interaction forces.}
	\label{fig:contact_definition}
\end{figure}

To mitigate this challenge, we define a new contact model. Occupied pixels are only considered for border forces, if they fall into a distance threshold and are closest in a quadrant relative to the agent's gazing direction (Figure~\ref{sfig:contact_quadrants}). The quadrants are defined such that the first quadrant (I) is centered about the agent's gazing direction. In all four quadrants, the partial border forces $\mathbf{f}_{i}^{B,k}$ are computed and agglomerated into the total border force $\mathbf{f}_{i}^{B}$ (Figure~\ref{sfig:contact_forces}). The partial border forces are computed as an exponential force superimposed by a linear force. We observe, that the pure exponential force used by Farina et al.~\cite{FFG17} acts too slow to prevent agents from penetrating walls if configured according to Table~\ref{tab:data}. The linear force rises faster while the final force is typically lower. Hence, the agent is decelerated but may still approach the wall. In very close regions, the exponential force forces the agent away from the borders. As agents approach borders with lower velocity, they cannot skip the border within an update cycle, hence, penetration is prevented. We define the partial border force as
\begin{equation}
	\label{eq:contact}
	\begin{split}
		\mathbf{f}_{i}^{B,k} &= \phi^{B}(r_{i}-d_{i}^{B,k})\mathbf{n}_{i}^{B,k}\\
		&+\begin{cases}
			\mathbf{0} & \text{, } d_{i}^{B,k}>r_{i}\\
			\begin{split}
				&C^{s}\phi^{s}(d_{i}^{B,k})\mathbf{n}_{i}^{B,k}\\
				&+(1-C^{s})\phi^{s}(d_{i}^{B,k})v_{i,y}\mathbf{t}_{i}^{B,k}
			\end{split}
			& \text{, otherwise.}
		\end{cases}
	\end{split}
\end{equation}
Hereby, $r_{i}$ is the radius of the agent and
\begin{equation}
	d_{i}^{B,k}=||\mathbf{p}_{i} - \mathbf{p}_{k} -  \frac{\sqrt{2}}{2}w\mathbf{n}_{i}^{B,k}||_{2}
\end{equation}
is the distance from the agent to the center of the closest occupied pixel $\mathbf{p}_{k}$ considering half the pixel width $w$, simplified as circle with conservative radius $0.5\sqrt{2}w$. The contact direction is split into the normalized normal and normalized tangential direction vectors
\begin{subequations}
	\begin{gather}
		\mathbf{n}_{i}^{B,k} = 
		\begin{bmatrix}
			n_{i,x}^{B,k}\\ n_{i,y}^{B,k}
		\end{bmatrix}
		= \frac{\mathbf{p}_{i}-\mathbf{p}_{k}}{d_{i}^{B,k}}, \quad
		\mathbf{t}_{i}^{B,k} =
		\begin{bmatrix}
			- n_{i,y}^{B,k}\\ n_{i,x}^{B,k}
		\end{bmatrix},
		\tag{\addtocounter{equation}{1}\theequation,\addtocounter{equation}{1}\theequation}
	\end{gather}
\end{subequations}
respectively. Note that the tangential part in Equation~\ref{eq:contact} scales with the agent's orthogonal velocity $v_{i,y}$. The border potential is defined as
\begin{equation}
	\label{eq:border_potential}
	\phi^{B}(r_{i}-d_{i}^{B,k}) = \phi_{0}^{B}\cdot \exp{(\frac{r_{i}-d_{i}^{B,k}}{C^{B}})}.
\end{equation}
The soft potential is defined as
\begin{equation}
	\phi^{s}(d_{i}^{B,k}) =
	\begin{cases}
		0 & \text{, } d_{i}^{B,k}> r_{i}\\
		\frac{r_{i}-d_{i}^{B,k}}{r_{i}-d_{min}^{B}}\phi_{0}^{s} & \text{, } d_{i}^{B,k}\le d_{min}^{B}\\
		\phi_{0}^{s} & \text{, otherwise,}
	\end{cases}
\end{equation}
where $d_{min}^{B}<r_{i}$ and $\phi_{0}^{s}\gg0$. The parameter configuration for the contact model is listed in Table~\ref{tab:contact_params}.
\begin{table}[t!]
	\caption{Parameter configuration of the contact model.}
	\centering
	\begin{tabular}{c c | c | c}
		\hline
		\textbf{Parameter} & \textbf{Description} & \textbf{Range} & \textbf{Value}\\
		\hline
		$C^{s}$        & Directional weight      & $[0,1]$      & $0.5$\\
		$\phi_{0}^{B}$ & Base potential (border) & $(0,\infty)$ & $11$\\
		$C^{B}$        & Scaling factor (border) & $(0,\infty)$ & $0.2$\\
		$\phi_{0}^{s}$ & Base potential (soft)   & $(0,\infty)$ & $1200$\\
		$d_{min}^{B}$  & Inflation               & $[0,r_{i})$  & $0$
	\end{tabular}
	\label{tab:contact_params}
\end{table}

\subsection{Further Extensions of Standard HSFM}
In addition to the already discussed alterations, we use certain features of state-of-the-art models. Firstly, we use the group cohesion proposed by Farina et al.~\cite{FFG17} to better depict squad behavior. We use different configurations of the potential in agent-agent interaction, which are interchanged if the agents are part of the same or different squads. The repulsion of squad mates is lower than in inter-squad interaction. Secondly, we model agent-agent and agent-border forces with the Elliptical Specification~(ES)~II~\cite{JHS07}. We experimentally compared ES~I~\cite{HM95}, ES~II, and Circular Specification~\cite{HFV00}. We observed that the ES~II works best with the new contact definition, resulting in less oscillation, i.e., better damping behavior, after first contact with borders.

\section{Validation}
\label{sec:validation}
\begin{figure}
	\centering
	\includegraphics[width=.47\textwidth]{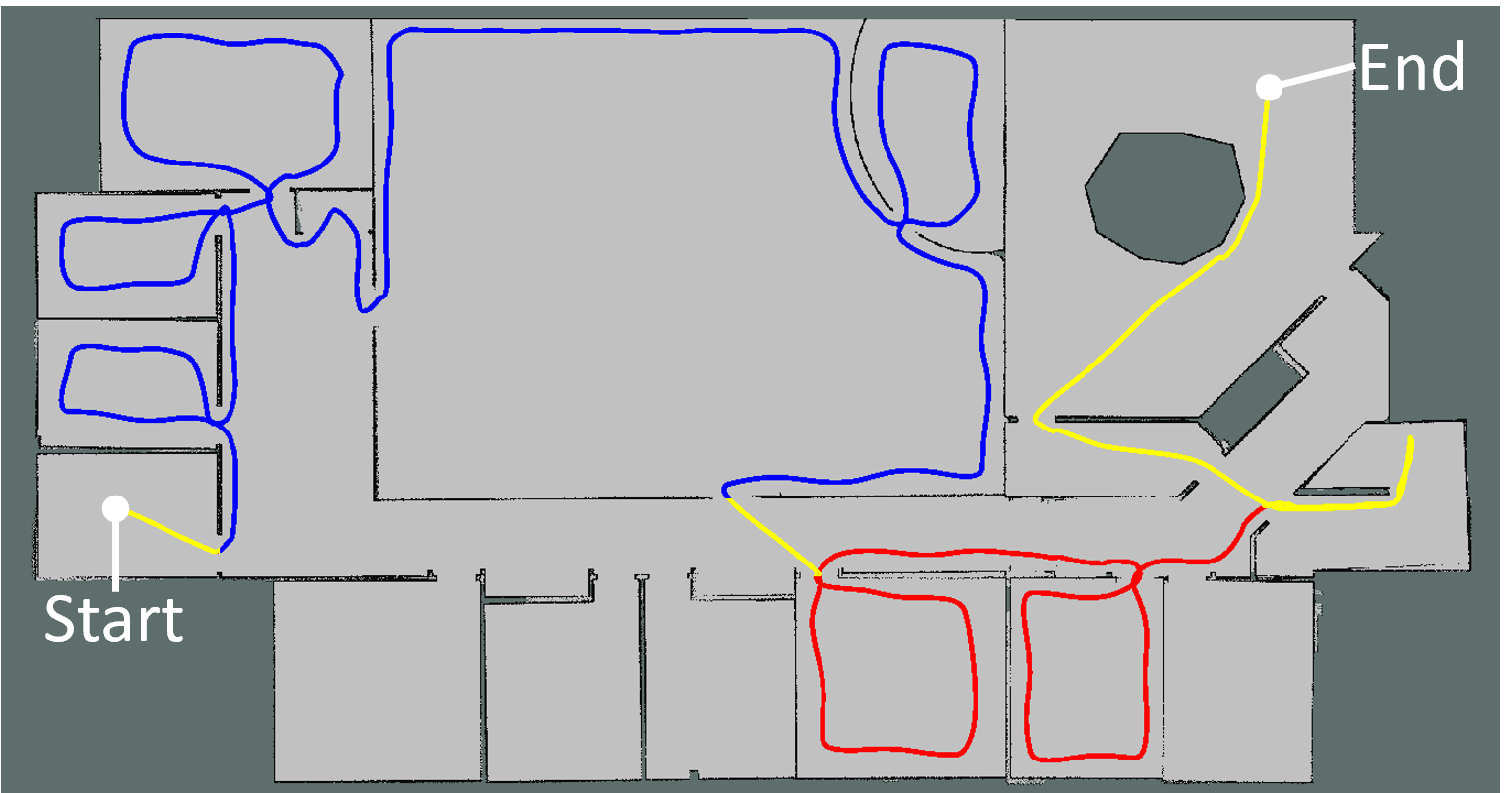}
	\setlength{\belowcaptionskip}{-15pt}
	\caption{Simulation of a single squad. Displayed is the path of a single agent (yellow: free, blue: LHR, red: RHR).}
	\label{fig:results_path}
\end{figure}

Test hardware consists of Intel i7-7700HQ and $8$Gb RAM. The modified HSFM is solved with a Runge-Kutta solver using the Dormand-Prince method of fifth order~\cite{CMR90} (\emph{runge\_kutta\_dopri5} in \emph{boost}) and a step size of $0.06$sec. Computational times for single squad traversal are listed in Table~\ref{tab:results_compute} ($1000$ samples per test case). A squad consists of three agents, which is common for German fire brigades. With the low computational times, the robot employing the proposed methodology can generate squad trajectories in real or near-real time, depending on the length of the predicted path. Figure~\ref{fig:results_path} depicts the paths of a single squad traversing the building with varying motion tactics. Given the parameter configuration in Table~\ref{tab:data}, the motion prediction generates feasible results. Hence, we conclude that the method is well suited for deployment in collaborative rescue applications. 
\begin{table}[t]
	\caption{Characteristic measures of single squad traversal.}
	\centering
	\begin{tabular}{r | c c | c c}
		\hline
		\textbf{Tactic} & \multicolumn{2}{c|}{\textbf{Path Length} (in m)} & \multicolumn{2}{c}{\textbf{Simulation Time} (in ms)}\\
		~ & Mean ($\mu_{d}$) & Variance ($s_{d}^2$) & Mean ($\mu_{t}$) & Variance ($s_{t}^2$)\\
		\hline
		Free & $8.835$  & $0.156$ & $21.154$ & $2.973$\\
        ~    & $48.365$ & $0.155$ & $105.937$ & $18.528$\\
		\hline
		LHR & $11.568$  & $0.008$ & $78.044$ & $30.453$\\
		~   & $108.445$ & $0.818$ & $842.665$ & $634.517$\\
		\hline
		RHR & $14.787$  & $0.021$ & $93.298$ & $27.704$\\
		~   & $65.981$  & $0.163$ & $482.274$ & $201.914$\\
		\hline
	\end{tabular}
	\label{tab:results_compute}
\end{table}

\section{Conclusions}
\label{sec:conclusions}
In this work, we proposed a novel motion prediction pipeline for SAR. The method combines optimal path planning based on three specialized types of graphs with a modification of the HSFM~\cite{FFG17}. The graphs model the tactics applied by firefighters in SAR, while the HSFM uses the paths to generate agent-individual motion trajectories. We modified the agent-border interaction by introducing a soft contact to allow traversal closer to borders. The models are configured with a novel data set. Finally, we showed that the proposed pipeline generates feasible trajectories in SAR and is computed fast depending on map size and agent count.

\addtolength{\textheight}{-12cm}  




\section*{ACKNOWLEDGMENT}
We like to thank the IdF NRW for providing data and Sebastian D\"{o}bler, Onur Akin, Lukas Clasen, and Till Waldermann for their support in data collection and implementation.

\bibliographystyle{IEEEtran}
\bibliography{IEEEabrv,references}

\end{document}